\documentclass[letterpaper, 10 pt, conference]{ieeeconf}  

\IEEEoverridecommandlockouts                              

\usepackage{graphicx} 

\usepackage{amsmath} 
\usepackage{amssymb}  
\usepackage{comment}
\usepackage{siunitx}
\usepackage{url}

\newcommand{\argmin}{\mathop{\rm arg~min}\limits}

\title{\LARGE \bf
	A Hand Motion-guided Articulation and Segmentation Estimation
}

\author{Richard Sahala Hartanto, Ryoichi Ishikawa, Menandro Roxas, Takeshi Oishi
	\thanks{Richard Sahala Hartanto, Ryoichi Ishikawa, Menandro Roxas and Takeshi Oishi are with Institute of Industrial Science, The University of Tokyo, Japan
		{\tt\small \{richard, ishikawa, roxas, oishi\}@cvl.iis.u-tokyo.ac.jp}}%
}

\begin{document}

\maketitle
\thispagestyle{empty}
\pagestyle{empty}

\section*{Abstract}

In this paper, we present a method for simultaneous articulation model estimation and segmentation of an articulated object in RGB-D images using human hand motion.
Our method uses the hand motion in the processes of the initial articulation model estimation, ICP-based model parameter optimization, and region selection of the target object.
The hand motion gives an initial guess of the articulation model: prismatic or revolute joint.
The method estimates the joint parameters by aligning the RGB-D images with the constraint of the hand motion.
Finally, the target regions are selected from the cluster regions which move symmetrically along with the articulation model.
Our experimental results show the robustness of the proposed method for the various objects.

\section{Introduction}
Understanding the articulation of objects is crucial for robots to manipulate various functional objects.
The robot needs to learn how to use the tools that humans use at home or in the workplace.
For that, the robot needs to understand movement patterns and ranges in 3D space, the articulation model, 
of these objects to manipulate them correctly and to avoid damages of the object, environment, and robot itself.

Most of the previous works on articulation estimation utilize visual input: an RGB or RGB-D image sequence.
The image-based method tracks feature points of a moving object in a 2D space \cite{ross2010learning,Jacquet_2013_CVPR,Pauwels_2014_CVPR} or a 3D space \cite{huang2012occlusion,pillai14}.
If depth information is given, 3D geometric features are useful \cite{pekelny2008articulated,tzionas2016reconstructing}.
Recently, Deep Neural Network-based approaches such as CNN for functional estimation \cite{chaudhary2018predicting} or a verbal-guided approach \cite{daniele2020multiview} have also been proposed.
Aside from visually supervised methods, the interactive approach using a robot arm is an alternative research direction  \cite{hausman2015active}, 
but we consider the visual supervision for general-purposes.

In addition to the articulation model estimation, the functional segmentation of the object is also important.
One of the popular approaches is a structure-driven method that assumes the target object is composed of rectangular planes \cite{sturm2010vision}.
The CNN-based approach is also successful in the segmentation process \cite{chaudhary2018predicting}.
Another stream is more closely linked to the articulation, {\it i.e}., tracking and clustering movements.
The point clouds that belong to a functional segment move along with the same articulation model.
Therefore, the segmentation is performed by tracking 2D feature points \cite{ross2010learning,fayad2011automated} or 3D structural feature points \cite{kim2016simultaneous,lu2018unsupervised} and by clustering the motions of the points.

\begin{figure}[t]
	\begin{center}
	\includegraphics[width=\linewidth]{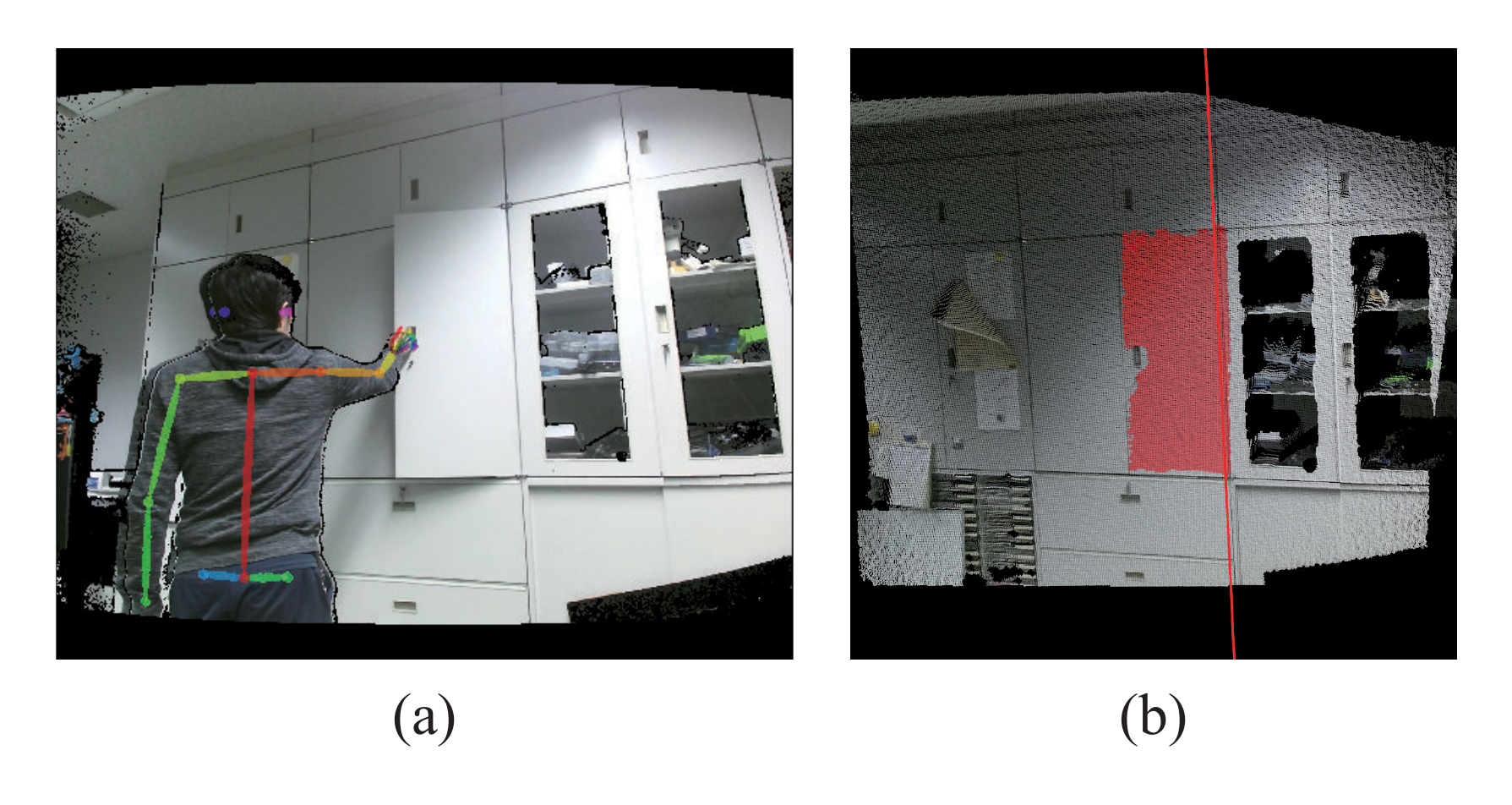}
	\caption{Articulation and segmentation estimation from RGBD images with using human's hand movement. (a) input RGB-D image sequence with human pose detection by OpenPose \cite{cao2018openpose}. (b) Articulation and segmentation estimation results.}
	\label{fig:intro}
	\end{center}
\end{figure}

Some problems with conventional methods are the dependency on textures and distinctive shapes of the object.
Since functional objects are usually textureless and symmetric, these problems are highlighted especially when using noisy depth information from optical sensors. In short, robustly tracking the target regions in 3D space is no easy task.

We consider using human motion while manipulating the articulated object.
The human interacts with the target object, and the contact point moves according to the articulation model. 
Recent CNN development makes it easy and accurate to detect human motion from RGB or RGB-D images \cite{he2017mask,Schroder_2017_CVPR_Workshops,cao2018openpose}, 
even in situations with partial occlusions.
We can easily observe the human motion during manipulation and the target regions at the same time.
\begin{figure*}[t]
	\begin{center}
	\includegraphics[width=.9\linewidth]{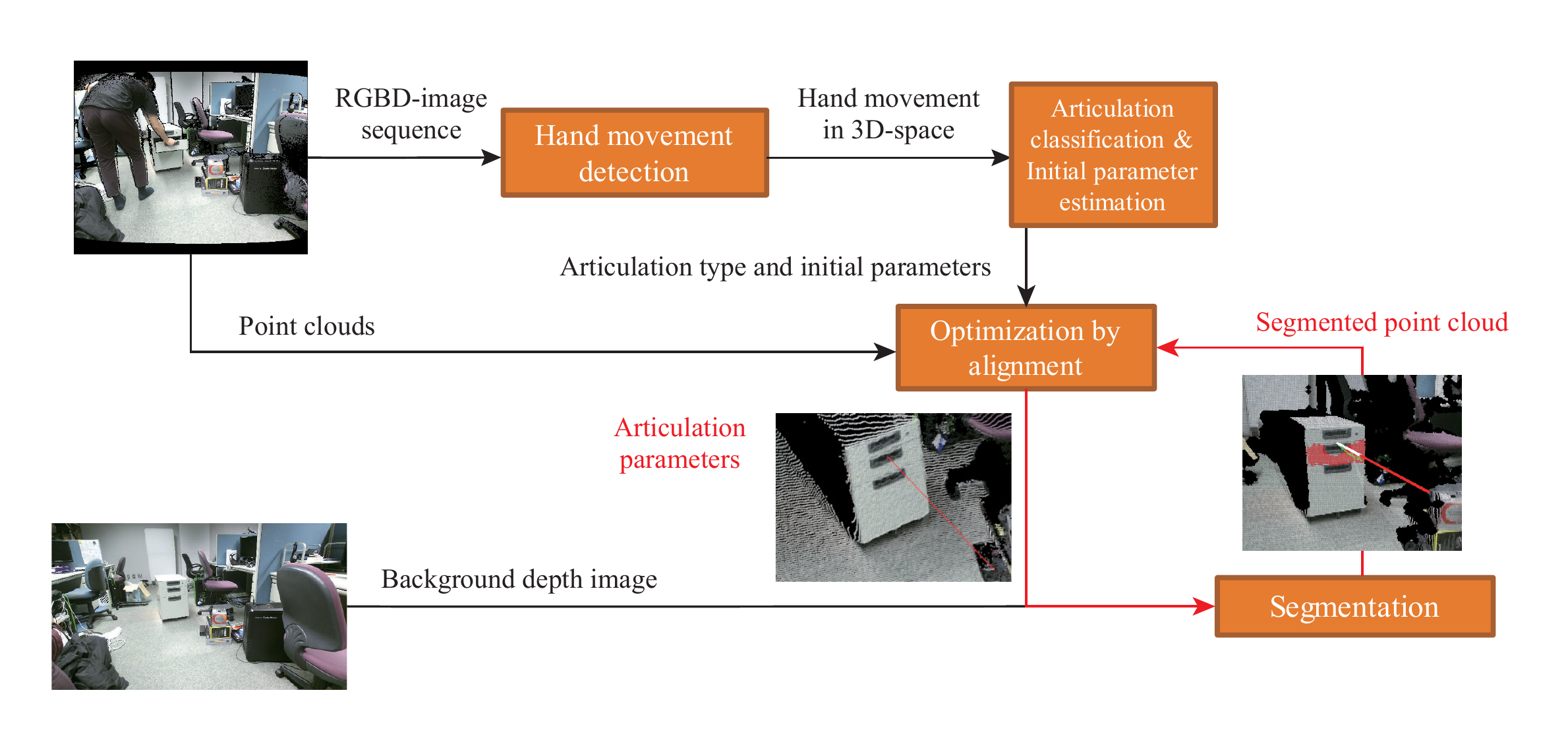}
	\caption{Overview of our method. The red letter indicates the outputs of our method. The loop indicated by the red arrow is the iterative refinement process.}
	\label{fig:overview}
	\end{center}
\end{figure*}

In this paper, we propose a method of hand motion-guided articulation and segmentation estimation from RGB-D images.
The inputs are the RGB-D images of a background scene and a sequence of manipulation (Fig.~\ref{fig:intro}~(a)).
The outputs are the articulation model and segmentation of the target object (Fig.~\ref{fig:intro}~(b)).
We detect hand motion using a CNN-based method \cite{cao2018openpose} and utilize it for initial estimation of the articulation model, weight and constraint for depth image alignment, and region selection. 
The proposed method does not primarily depend on the object shape and textures since we estimate the initial model from the human body information, which is characteristic and detectable using modern CNN-based methods.
The experimental results show the robustness of our method for various objects.
The code is released as an open-source\footnote{\url{https://github.com/cln515/Articulation-Estimation}}.

\section{Overview and notation}

\subsection{Overview}

We use an RGB-D video that captures a scene of manipulating the articulated object and an RGB-D image of background. Figure~\ref{fig:overview} shows an overview of the proposed method.

We assume the following conditions:
\begin{itemize}
	\item Input RGB-D video includes only a manipulation scene, and the first frame is the start of manipulation.
	\item Only one person is in the RGB-D video.
	\item It is known which hand is used.
	\item Articulation type is either prismatic or revolute as described in \ref{sec:articulation_model}.
\end{itemize}

First, we estimate the hand motion in 3D space from RGB images and depth images using CNN-based semantic segmentation and human pose estimation method. 
Next, the articulation type and the initial parameters are estimated from the hand motion. 
The alignment is applied to the point clouds derived from the depth images to optimize the articulation parameters according to the estimated articulation type.
The segmentation is performed by extracting points which are symmetric and move according to the articulation model.
The alignment using the segmentation result refines the parameters. The refinement process is shown as the red arrows in Fig.~\ref{fig:overview}. (In our experiments, we iterate the refinement process twice)

\subsection{Articulation model}
\label{sec:articulation_model}
\begin{figure}[th]
	\begin{center}
	\includegraphics[width=\linewidth]{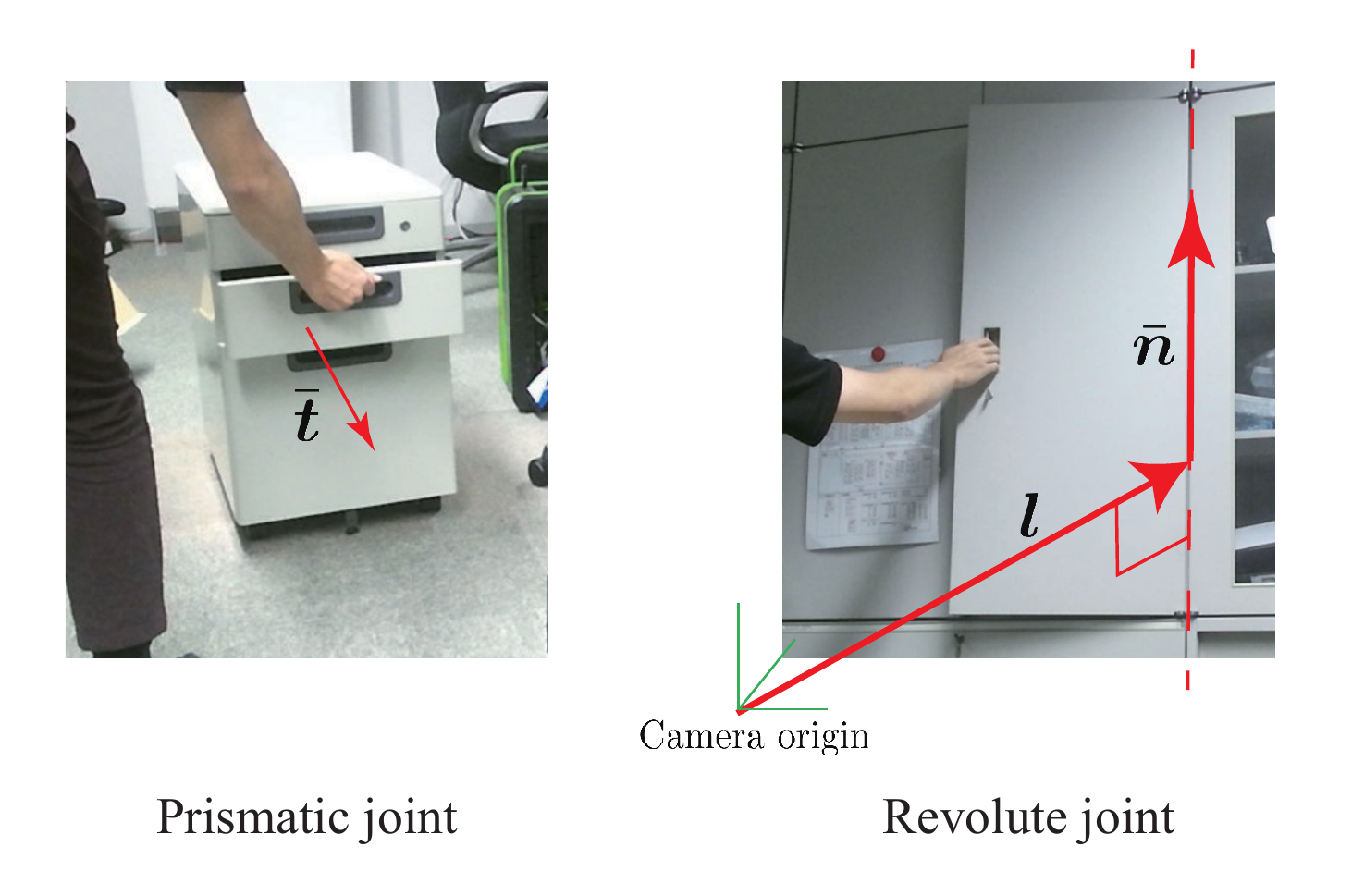}
	\caption{Example of a prismatic joint and a revolute joint. The prismatic joint object transformed along the direction $\bar{\boldsymbol{t}}$. The revolute joint object rotates around the axis where going through $\boldsymbol{l}$ in the direction $\bar{\boldsymbol{n}}$}
	\label{fig:articulation}
	\end{center}
\end{figure}
We treat an object as either containing a prismatic joint or a revolute joint. These joints are the basis of most objects designed for humans. We describe how we express these joints as follows.

\subsubsection{Prismatic joint}
\label{secpris}
When a rigid object is linked to a sliding linkage, in which we call the prismatic joint, the movement of the object is restricted to make only a linear sliding movement relative to the linkage (See the left of Fig.~\ref{fig:articulation}).
The prismatic joint has only one direction to move in 3D space.
We express the prismatic joint as 2-DoF normalized $3 \times 1$ vector $\bar{\boldsymbol{t}}$ indicating moving direction.
The amount of movement of a prismatic object is represented by $a$.
The unit of $a$ can be any but is a meter in the experiment. 
An object moving along the prismatic joint is transformed by the following $4 \times 4$ matrix;
\begin{align}
\left( 
\begin{array}{cc}
\boldsymbol{I}_{3\times3} & a\bar{\boldsymbol{t}} \\
\boldsymbol{0}_{1\times3} & 0
\end{array}	
\right).
\end{align}
where, $\boldsymbol{I}_{3\times3}$ is $3\times3$ identity matrix.

\subsubsection{Revolute joint}
\label{secrev}
When a rigid object is linked to a rotating linkage, in which we call the revolute joint, the movement of the object is restricted to only rotation around an axis with a certain movement range relative to the linkage (See the right of Fig.~\ref{fig:articulation}). 
The revolute joint can be expressed as a line in 3D space.
We express the revolute joint as a total of 4-DoF parameters: 2-DoF unit vector $\bar{\boldsymbol{n}}$ of the rotation axis and 2-DoF vector $\boldsymbol{l}$ where the rotation axis passes through and holds $\boldsymbol{n}\cdot\boldsymbol{l}=0$.
The amount of movement of a revolute object is represented by $\theta$, and the unit is the radian. 
An object moving along the revolute joint is transformed by the following $4 \times 4$ matrix;
\begin{align}
\left( 
\begin{array}{cc}
\boldsymbol{R}(\theta, \bar{\boldsymbol{n}}) & \boldsymbol{l}-\boldsymbol{R}(\theta, \bar{\boldsymbol{n}})\boldsymbol{l} \\
\boldsymbol{0}_{1\times3} & 0
\end{array}	
\right).
\end{align}
where, $\boldsymbol{R}(\theta, \bar{\boldsymbol{n}})$ is $3\times3$ rotation matrix indicating rotation by $\theta$ around $\bar{\boldsymbol{n}}$:
\begin{align}
	\boldsymbol{R}(\theta, \bar{\boldsymbol{n}})=  \cos \theta \boldsymbol{I} + (1-\cos \theta) \bar{\boldsymbol{n}} \bar{\boldsymbol{n}}^\top + \sin \theta [\bar{\boldsymbol{n}}]_\times,
\end{align}
where $[\cdot]_\times$ is $3 \times 3$ skew symmetric matrix.

\section{Methodlogy}
\subsection{Initial articulation estimation}
We recognize the articulation type and estimate the initial articulation parameters from the hand motion during manipulation in the 3D space. To detect the hand position in RGB images, we use hand keypoint detection \cite{simon2017hand} implemented in OpenPose \cite{cao2018openpose}.
The hand keypoint detection offers a detailed hand pose as a set of 2D points and it needs a bounding box around the hand area.
The bounding box is offered from OpenPose when the entire body is pictured or CNN-based hand detection implementation \cite{Dibia2017} trained using the EgoHands dataset \cite{Bambach_2015_ICCV}. 
After obtaining a hand pose in 2D space, we compute the 3D points of the hand points using the depth map and compute the 3D centroid of the hand.

After estimating the hand motion, we fit a circle to the motion path for recognizing the articulation type. We use RANSAC and a non-linear optimization method to fit the circle to the noisy points. Counting the number of inlier points is not practical since a straight line can be considered as a part of a large circle. 
Therefore, we use the range of the movement angle for type recognition. If the range is smaller than a predetermined threshold value (we set the value to $30$ degrees), we recognize the target object to be prismatic and perform the line fitting.

We estimate the initial articulation parameters from the result of the fitting.
The line fitting computes 4-DoF parameters as explained in Sec.~\ref{secpris}.
The circle parameters are 6-DoF parameters consisting of the radius $r$ (1-DoF), center point $\boldsymbol{c}$ (3-DoF) and a normal vector of the circle $\bar{\boldsymbol{n}}$ (2-DoF). 
In the case of the prismatic joint, the direction of the fitting line becomes the direction $\bar{\boldsymbol{t}}$.
In the case of the revolute joint, the axis passes through $\boldsymbol{c}$ in the direction $\bar{\boldsymbol{n}}$.
Since $\boldsymbol{l}\cdot\bar{\boldsymbol{n}}=0$ holds, the position parameter of the axis $\boldsymbol{l}$ is;
\begin{align}
	\boldsymbol{l} = \boldsymbol{c}-(\boldsymbol{c}\cdot\bar{\boldsymbol{n}})\bar{\boldsymbol{n}}.
\end{align}

\subsection{Articulation parameter optimization}
\label{sec:alignment}

The articulation parameters are optimized by aligning the point clouds of the manipulated object through the sequence.
Since the point cloud sequence includes both the static background and the dynamic objects, the alignment of the manipulated object needs to ignore the effect of the static background.
We assume that points nearby the detected hand are likely to belong to the manipulated object.
Therefore, we introduce a weighting scheme on 3D points using the hand position; we set the weighting parameter according to the distance from the hand to each point.

The parameters in the optimization process are the articulation parameters and the amount of movement of object between each frame.
For simplicity, let $\boldsymbol{J}$ be the articulation parameters and $m_i$ be the amount of movement of the manipulated object in $i$-th frame.
In the case of prismatic joint, $\boldsymbol{J}$ is the moving direction $\bar{\boldsymbol{t}}$, and $m_i$ is the distance parameter $a$ in Sec.~\ref{secpris}.
In the case of revolute joint, $\boldsymbol{J}$ is the rotation axis $\bar{\boldsymbol{n}}$ and $\boldsymbol{l}$, and $m_i$ is the amount of rotation $\theta$ in Sec.~\ref{secrev}.

We align the point clouds of multiple frames simultaneously by a constrained ICP. Let $\boldsymbol{p}_{k}$ be $k$-th points in $i$-th frame and $\boldsymbol{q}_{k}$ be the closest point of $\boldsymbol{p}_{k}$ in $j$-th frame.  
Considering constrained ICP alignment with only two frames, $i$-th and $j$-th, the error is described as follows:
\begin{align}
	\begin{split}
		\varepsilon_{i,j} &= \sum_{k} w_{k}w'_{k} |((\boldsymbol{R}_i(\boldsymbol{J}, m_i) \boldsymbol{p}_{k})+\boldsymbol{t}_i(\boldsymbol{J}, m_i)) -\\
		&((\boldsymbol{R}_j(\boldsymbol{J}, m_j) \boldsymbol{q}_{k})+\boldsymbol{t}_j(\boldsymbol{J}, m_j))|,
	\end{split}
\end{align}
where $\boldsymbol{R}_i$ is the rotation matrix constrained by the joint parameters $\boldsymbol{J}$ with the amount of movement of $i$-th frame $m_i$, and $\boldsymbol{t}_i$ is the translation vector as well.
$w_{k}, w'_{k}$ are the weighting parameters according to the distance from the hand as described above and we define them as follows:
\begin{align}
	\label{eq:weight}
w_{k} =  \begin{cases}
	\left(\frac{1}{C + |\boldsymbol{p}_k - \boldsymbol{h}_i|}\right)^2 & (hand\:position\:is\:detected)\\
	1 & (otherwise)
\end{cases}\\
w'_{k} =  \begin{cases}
	\left(\frac{1}{C + |\boldsymbol{q}_{k} - \boldsymbol{h}_j|}\right)^2 & (hand\:position\:is\:detected)\\
	1 & (otherwise)
\end{cases}
\end{align}
where $\boldsymbol{h}_i$ is hand position in $i$-th frame and $C$ is a constant value empirically determined. The optimal parameters $\boldsymbol{\hat{J}}, (\hat{m}_1\dots \hat{m}_N)$ are obtained by simultaneously minimizing the error as follows:
\begin{align}
	\label{eq:icp}
	\boldsymbol{\hat{J}}, (\hat{m}_1\dots \hat{m}_N) = \argmin_{\boldsymbol{J}, (m_1\dots m_N)} \sum_{i,j} \varepsilon_{i,j},
\end{align}
where $N$ is the number of frames.

\subsection{Segmentation}
\label{sec:seg}
\begin{figure}[th]
	\begin{center}
	\includegraphics[width=\linewidth]{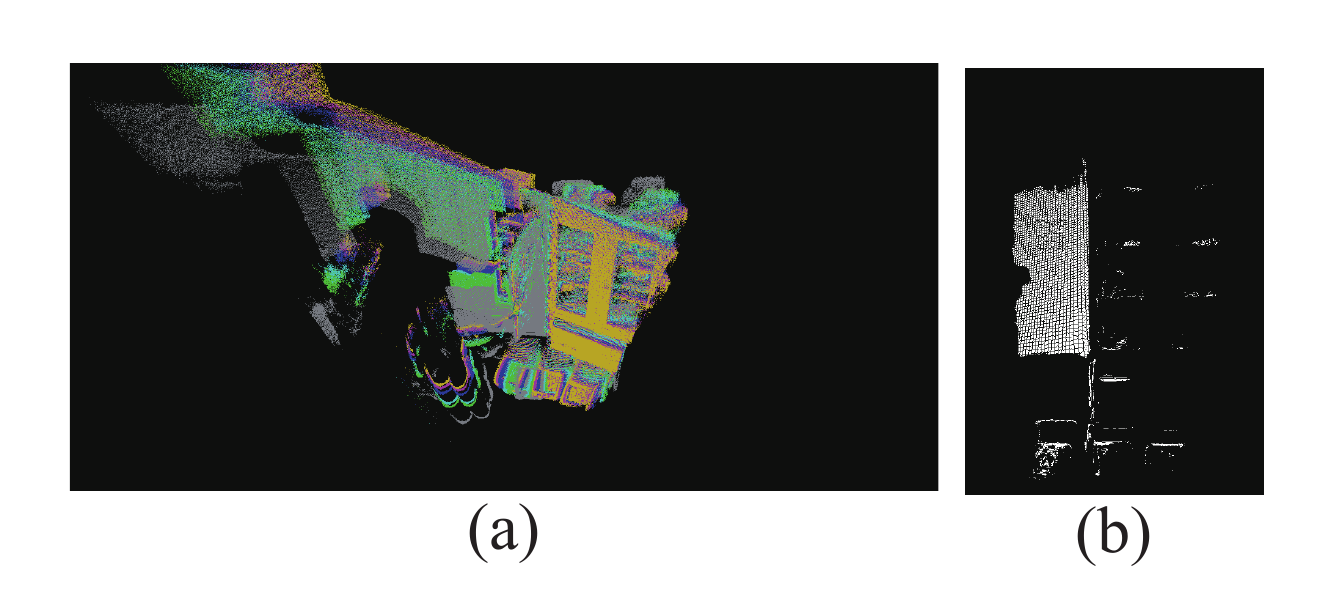}
	\caption{(a) Point clouds after the alignment of the manipulated object. (b) Extracted symmetric objects.}
	\label{fig:segm}
	\end{center}
\end{figure}

We segment the points in the sequence of the point clouds into the manipulated object and the background. After aligning the point clouds as described in the previous section, the areas that move symmetrically about the articulation axis overlap each other, as shown in Fig.~\ref{fig:segm}~(a). We first extract the points in these areas as the potential regions of the manipulated object, as shown in Fig.~\ref{fig:segm}~(b). Next, since the extracted points include ambiguous regions, we select the regions of the manipulated object by using a clustering method and the hand position by assuming that the object is connected to the hand that manipulates it.

\subsubsection{Initial points extraction}
We consider that the points of the manipulated object move symmetrically according to the articulation model through the frames. For example, in the case of the prismatic drawer, the points on the drawer's surface move in parallel along the direction of movement. In the case of the revolute drawer, such as a door, the points move symmetrically about the axis of rotation. Such areas overlap with each other in the sequential frames after alignment in the previous section. In other words, the error between the corresponding points is small in these areas.

We obtain the points in the overlapping areas using the alignment error. If the distance from a point in each frame to the closest point in the background frame is less than a threshold value, we extract the point as the potential manipulated object's point. The threshold value depends on the sensor accuracy. In our experiments, after calculating distances from the background frame to each frame, we used the median distance and set the threshold to $5\:cm$ for the first time and $3\:cm$ for the subsequent iterations in the refinement process described later in Sec.~\ref{sec:refine_opt}.

\begin{figure}[t]
	\begin{center}
	\includegraphics[width=\linewidth]{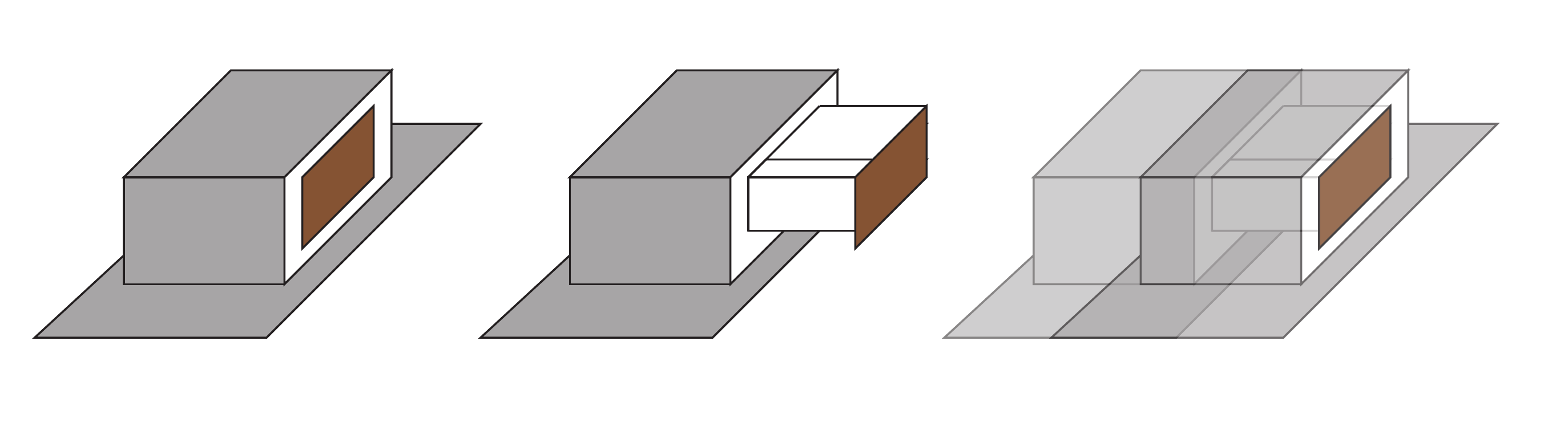}
	\caption{Prismatic object alignment. After aligning the brown area, gray areas are also detected as a symmetric area because of these overlap between before and after the movement.}
	\label{fig:pris_segm}
	\end{center}
\end{figure}
\begin{figure}[t]
	\begin{center}
	\includegraphics[width=\linewidth]{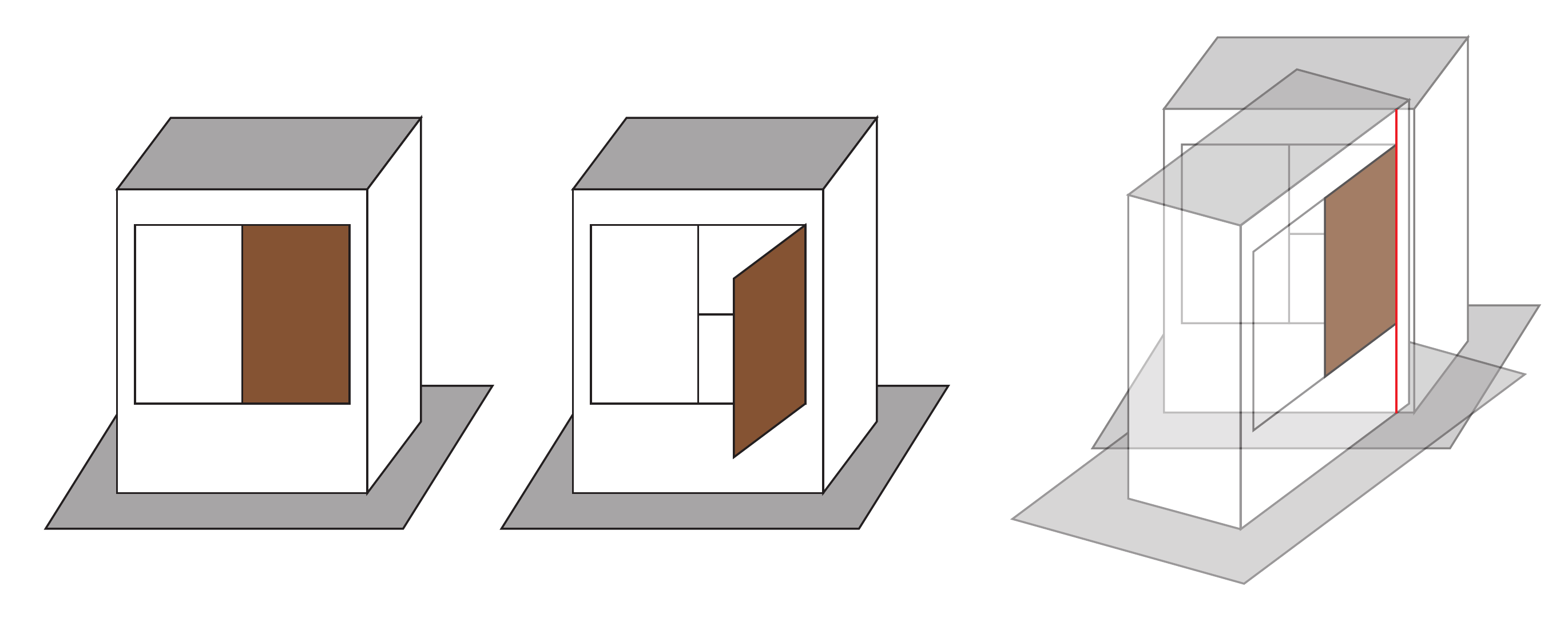}
	\caption{Revolute object alignment. After aligning the brown area, gray areas are also detected as a symmetric area because of these overlap between before and after the movement.}
	\label{fig:revo_segm}
	\end{center}
\end{figure}

\subsubsection{Region selection}
\begin{figure}[t]
	\begin{center}
	\includegraphics[width=.9\linewidth]{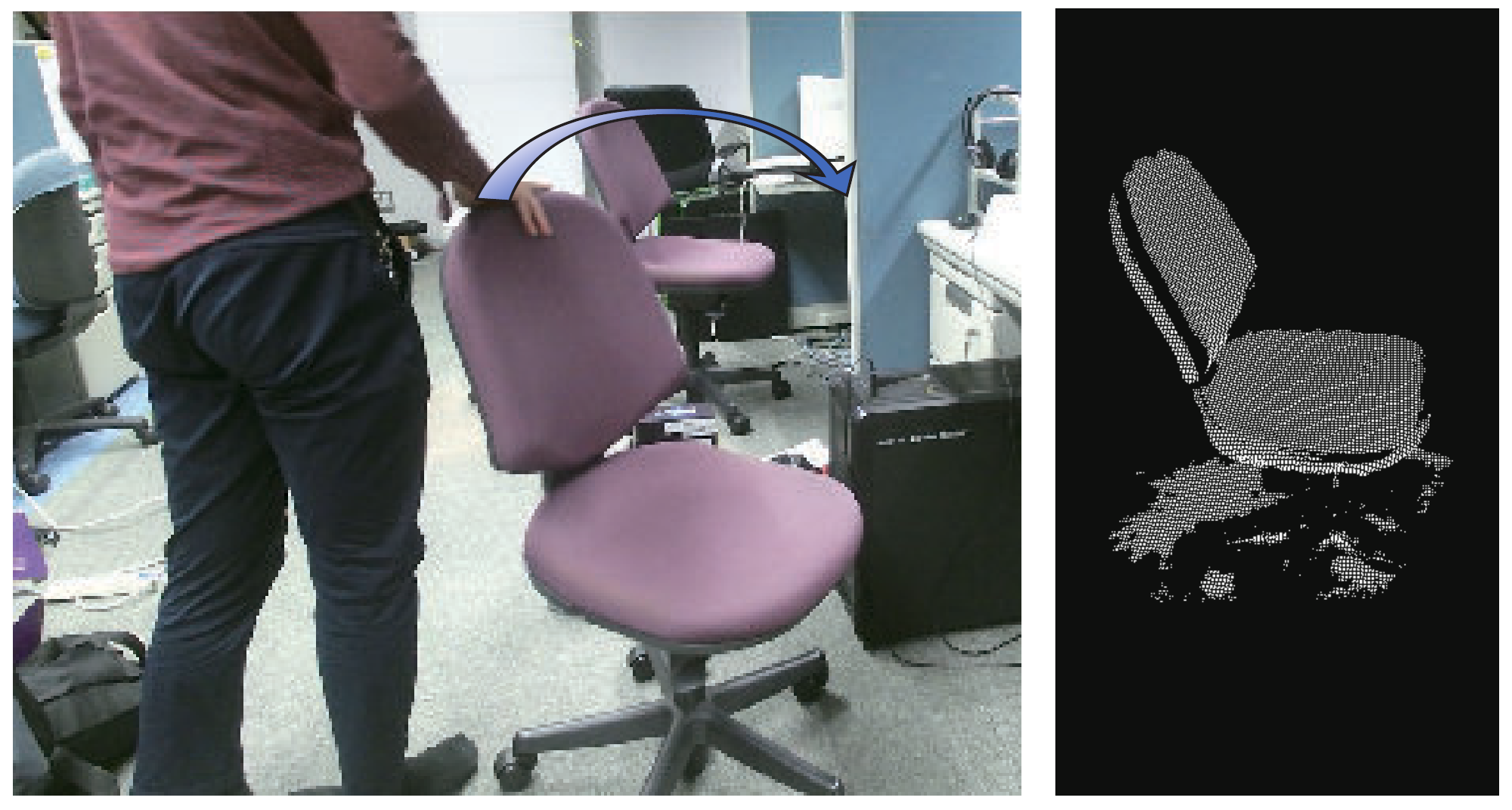}
	\caption{Example of symmetric areas of a revolute joint. A part of the floor is detected as a symmetric area due to its symmetric structure on the rotational axis of the chair.}
	\label{fig:revo_exs}
	\end{center}
\end{figure}

The extracted point cloud contains confidence and ambiguous regions.
The confidence region really moves according to the articulation model. On the other hand, the ambiguous region looks moving according to the model, but it is not easy to identify it to be background or target. For example, in the case of the prismatic joint, a region like a floor in Fig.~\ref{fig:pris_segm} whose surface normal is perpendicular to $\bar{\boldsymbol{t}}$ is ambiguous. In the case of the revolute joint, a region like a ceiling or a floor in Fig.~\ref{fig:revo_segm} that is symmetric about the rotation axis is also ambiguous. Figure~\ref{fig:revo_exs} shows an example of the symmetric area including ambiguous objects.

Therefore, we need to select points of the manipulated object from the initially extracted point cloud.
The confidence region can be identified by the surface normal and the articulation model. 
The surface normal of the point in the confidence region is the same direction to $\bar{\boldsymbol{t}}$ in the case of the prismatic joint.
In the case of the revolute joint, the surface normal moves along the tangential direction about the rotation axis.
We use the hand information again to select the points from ambiguous regions by assuming that the manipulated object connects to the hand. We apply Euclidian clustering \cite{rusu2010semantic} on the extracted point cloud and select the clusters nearby the hand position in the first frame with a threshold distance.

\subsection{Refinement with a hand soft constraint}
We refine the articulation parameters by aligning the segmented point clouds. Since the first alignment process in Sec.~\ref{sec:alignment} uses all points, including non-manipulated objects, the estimated parameters still have room for improvement in the accuracy. On the other hand, in the case of a feature-less object, the segmented point cloud does not have distinctive features for alignment. Therefore, we introduce a soft constraint by the hand in the refinement process.

\subsubsection{Geometric error term}
We define the geometric error by the distance of the corresponding points between the segmented points in the first frame and the corresponding points in $i$-th frame. The cost of the geometric error $c_{geo}$ is defined as the same with Eq.~\ref{eq:icp} with constant weights $w_k = w'_k = 1$ as follows:
\begin{align}
	\label{eq:geometric_error}
	c_{geo} = \sum_{j} \varepsilon_{0,j}
\end{align}
Note that $\varepsilon_{0,j}$ includes the error of only the segmented points.

\subsubsection{Hand soft constraint}
\begin{figure}[t]
	\begin{center}
	\includegraphics[width=\linewidth]{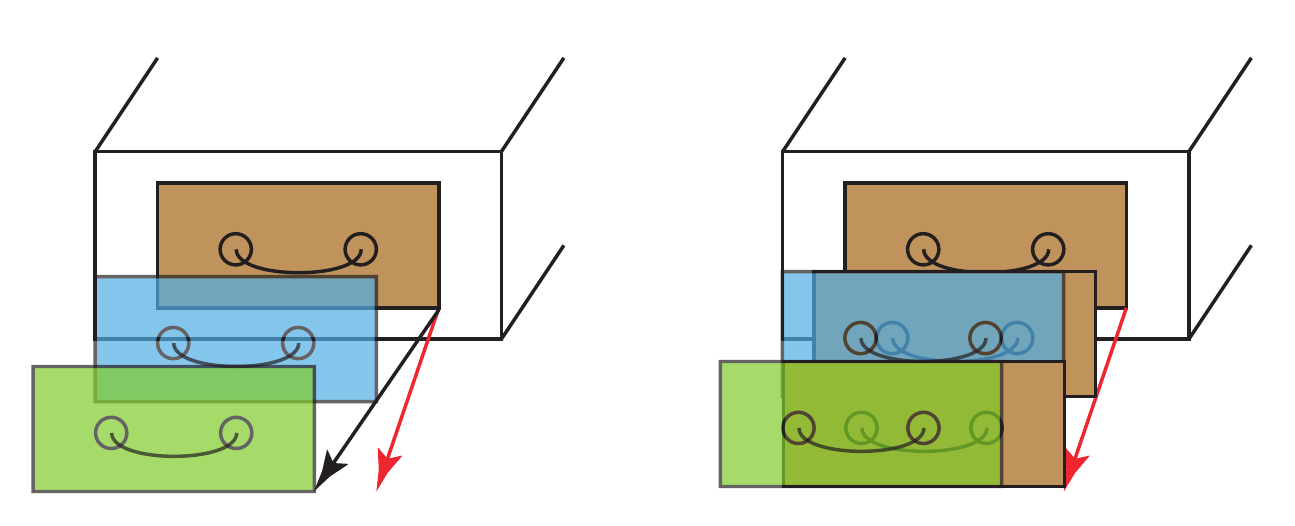}
	\caption{Alignment ambiguity in a flat prismatic object. Consider the case where the flat drawer moves along with the black arrow, however the estimated direction is the red. The alignment performs with even the wrong estimated direction 
	by treating the non-overlapped parts as outliers of alignment}
	\label{fig:hand_const}
	\end{center}
\end{figure}

We introduce soft constraint of hand to support the alignment of relatively feature-less surfaces and to eliminate the ambiguous regions. For example, in the case of prismatic objects as shown in Fig.~\ref{fig:hand_const}, the alignment error in the region where the surface is almost flat in the direction parallel to $\bar{\boldsymbol{t}}$ becomes small even the corresponding points are wrong by the sliding effect.

We use the 2D positions of the hand joints which are also given by the hand-keypoint detector.
The soft constraint of hand is the summation of the geometric error of each joint position.
The cost of soft constraint of hand $c_{hand}$ is,
\begin{align}
	c_{hand} = \sum_{l} \sum_{j} \alpha_{0,l}\alpha_{j,l} \left| \boldsymbol{h}_{0,l} - ((\boldsymbol{R}_j \boldsymbol{h}_{j,l})+\boldsymbol{t}_j)\right|,
\end{align}
where $\boldsymbol{h}_{j,l}$ is the position of $l$-th hand joint in $j$-th frame, and $\alpha_{j,l}$ is its confidence value.
The confidence value is also given by the detector.

\subsubsection{Optimization}
\label{sec:refine_opt}
Finally, the parameters are optimized by minimizing the joint cost described as follows:
\begin{align}
	\label{eq:opt}
	\boldsymbol{\hat{J}}, (\hat{m}_1\dots \hat{m}_N) = \argmin_{\boldsymbol{J}, (m_1\dots m_N)} \frac{1}{n} c_{geo} + \lambda c_{hand},
\end{align}
where $n$ is number of points in the segmented region, and $\lambda$ is the weight value which is empirically given to adjust the effect of the hand constraint.
After the parameter refinement, the segmentation result is also refined by the same process described Sec.~\ref{sec:seg}.

\section{Implementation}
The scanned person and apparent static objects are removed from point clouds to reduce the number of points for computational efficiency and improve the robustness. We apply RCNN-Mask \cite{matterport_maskrcnn_2017} to RGB images to find the region of the person and remove the points in the region.
The static objects can be removed by comparing the depth values in each pixel of the background frame and the manipulation sequence. 

We use a downsampled point cloud in the initial estimation process for computational efficiency.
We also limit the number of frames up to $15$ frames.
If the number of frames is higher than that, we sub-sample $15$ frames from the original sequence in which the hand positions are correctly detected.
We use kd-tree \cite{ann} for searching nearest neighbor points and Voxel Grid Filter to query in the downsampled kd-tree.
After the first segmentation, we use all points for the constrained ICP.

\begin{figure}[t]
	\begin{center}
	\includegraphics[width=\linewidth]{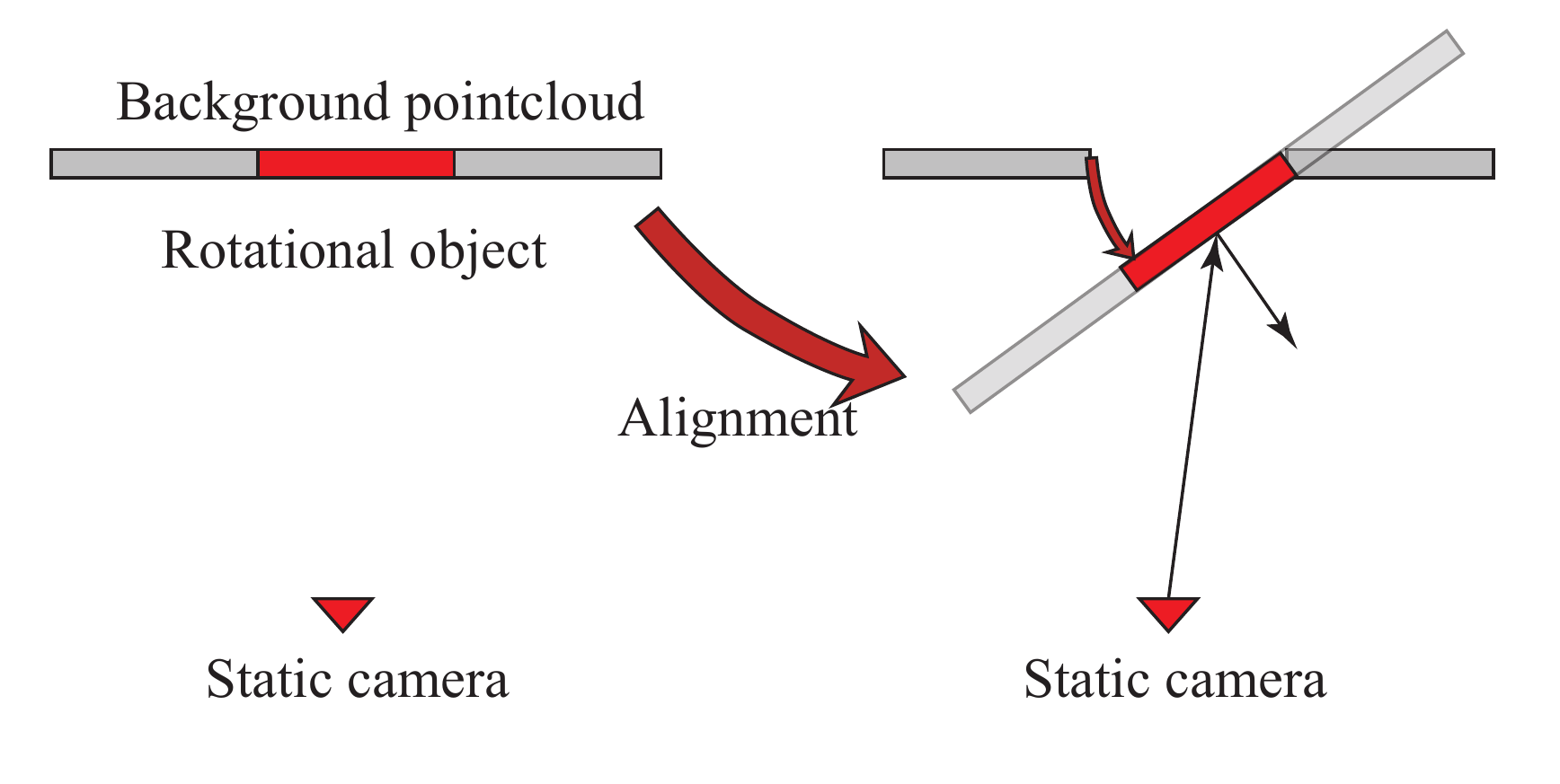}
	\caption{We remove points in a frame with a shallower angle of incidence from the camera for median computation of distance to the nearest points in each frame.}
	\label{fig:imple}
	\end{center}
\end{figure}

In the segmentation process, we also use surface normal information for filtering points out.
The accuracy of depth measurement gets worth when an incident angle of light is small.
We derive the symmetric region with filtering out the points in a frame with a smaller incident angle from the camera (See Fig~\ref{fig:imple}).

We also have several empirically determined values. We set $C=0.2$ in Eq.~\ref{eq:weight} and $\lambda=0.01$ in Eq.~\ref{eq:opt}. We used the implementations in \cite{pcl} for the Euclidean clustering for segmentation, Voxel Grid Filter, and kd-tree. We also used ceres-solver \cite{ceres-solver} for the non-linear optimization. 

\section{Experimental results}
We first validate the accuracy of the estimated articulation model.
We also demonstrate that the proposed method works well in various scenes.
We used Microsoft Kinect v2 \cite{kinect_site} as the RGB-D input device in all experiments. The length of the original input RGB-D sequences was around 10-40 frames. 

\subsection{Accuracy evaluation}
We used a flat prismatic object for validating the effect of soft hand constraint.
In addition to this, we demonstrate that the proposed method estimates the articulation model and segmentation of a revolute joint object accurately.

\subsubsection{Validation of hand soft constraint}
\begin{figure}[t]
	\begin{center}
	\includegraphics[width=\linewidth]{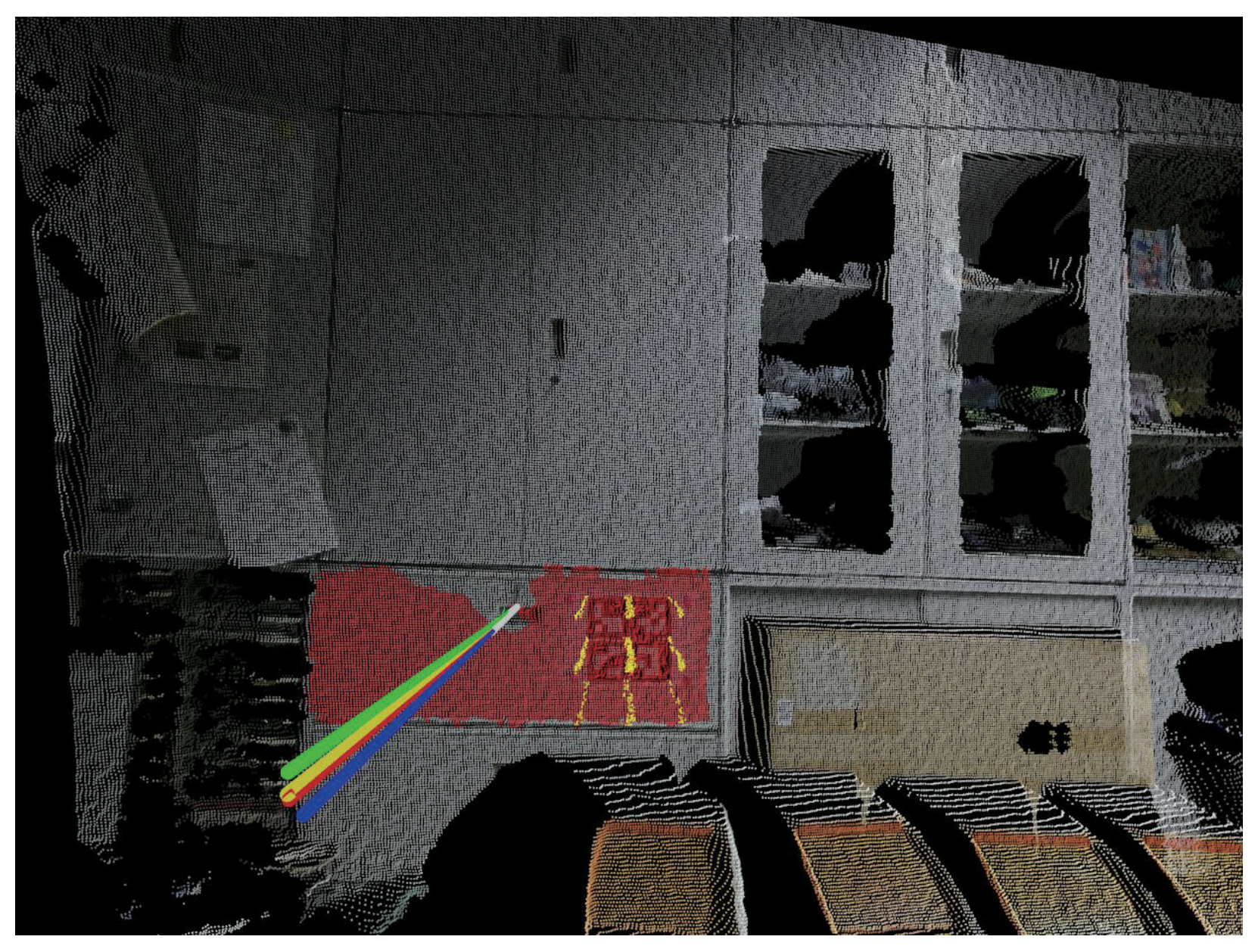}
	\caption{The articulation and segmentation estimation result with a flat drawer. (Green) Line fitting with the hand positions. (Yellow) Estimation using the trajectory of ArUco code corners, the corner trajectories are also shown as the yellow points. (Blue) The estimation result without hand soft constraint, (Red) The estimation result with hand soft constraint. Red points are the segmentation result with our method.}
	\label{fig:pris_ev}
	\end{center}
\end{figure}
\begin{figure}[t]
	\begin{center}
	\includegraphics[width=\linewidth]{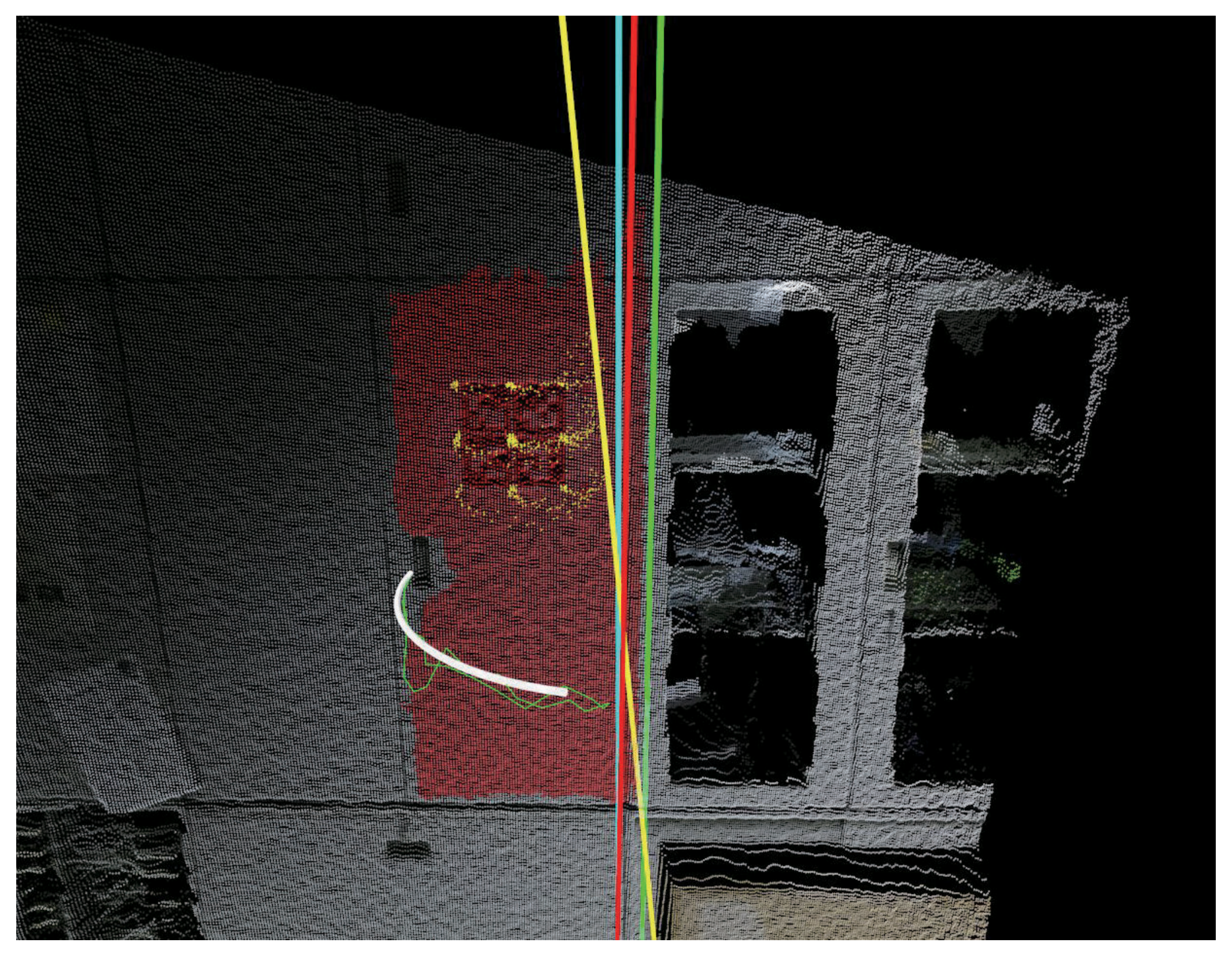}
	\caption{The articulation and segmentation estimation result with a shelf door. (Light blue) The parameters manually calculated from the hinge structure. (Green) Line fitting with the hand positions. (Yellow) Estimation using the trajectory of ArUco code corners, the corner trajectories are also shown as the yellow points. (Blue) The estimation result without hand soft constraint, (Red) The estimation result with hand soft constraint. Red points are the segmentation result with our method. (White) movement range}
	\label{fig:rev_ev}
	\end{center}
\end{figure}

\begin{figure*}[th]
	\begin{center}
	\includegraphics[width=\linewidth]{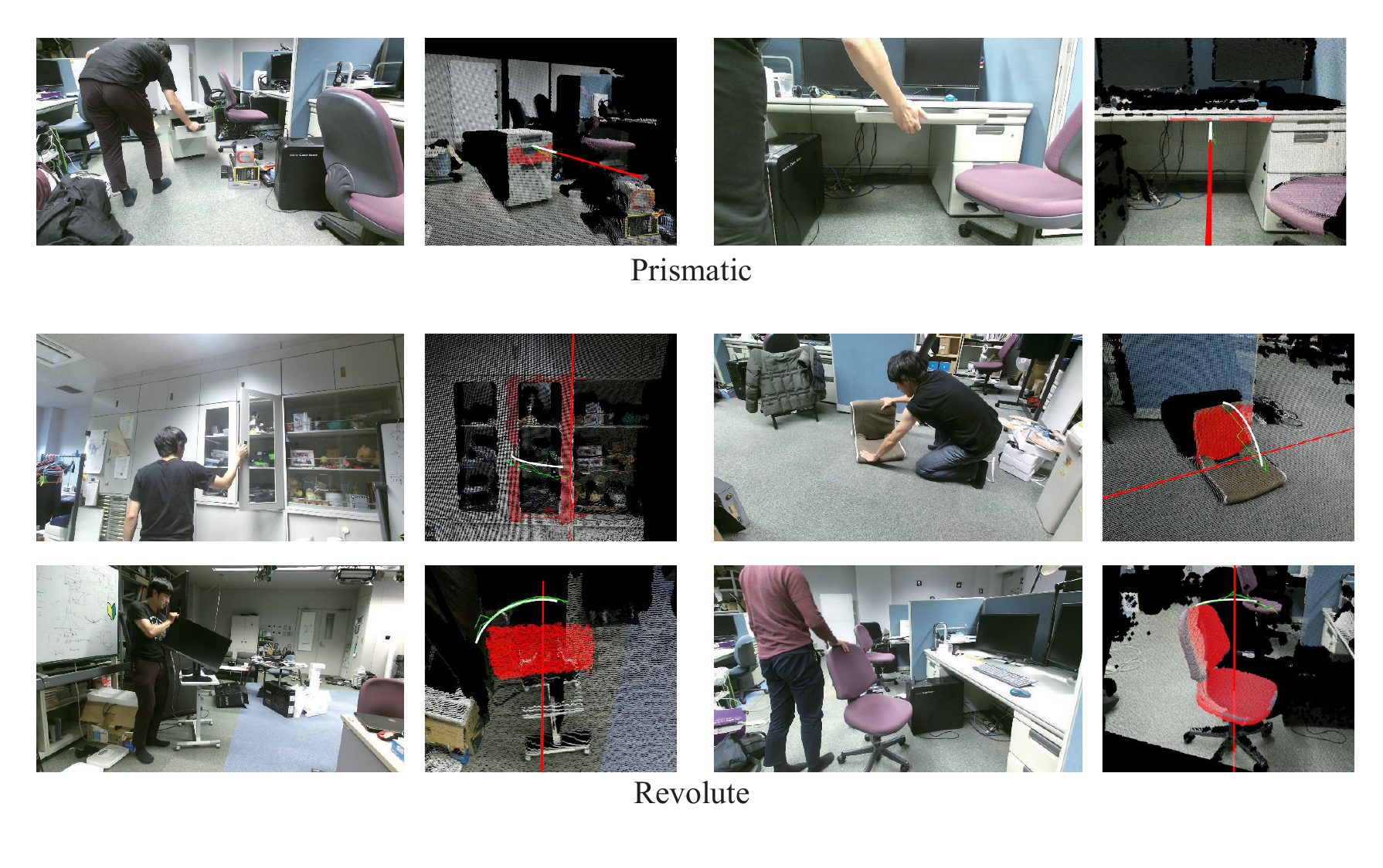}
	\caption{Other manipulation scenes and articulation and segmentation results. The articulation estimation results are shown as the red line. The segmentation results also shown as the red parts. Hand trajectory is also shown as the green polygonal line. The white lines indicate movement range in the video sequence. The top two are prismatic joint and the rest are revolute joint.}
	\label{fig:others}
	\end{center}
\end{figure*}

We used a flat drawer and pasted ArUco code on it for the comparative evaluation. 
Figure~\ref{fig:pris_ev} shows the estimation result of hand motion fitting (green), ArUco code (yellow), final estimation result with soft hand constraint (red), and without it (light blue).
Since the ArUco code detection works well, we can assume that the result of the ArUco code is a reference result in this case.
Comparing to the ArUco code result, the constrained ICP with hand motion is almost in the same direction, whereas the hand motion fitting and original ICP show an apparent error in the direction.
The result indicates that the human hand information works well for the articulation model estimation.

\subsubsection{Accuracy evaluation in revolute object}

Figure~\ref{fig:rev_ev} shows the result of the revolute articulation estimation.
In this case, we manually give the reference result by pointing out the hinges (right blue).
The estimation result of hand motion fitting (green), ArUco code (yellow) and final segmentation result with hand soft constraint (red) are shown in the figure.
The result by ArUco (yellow) has a more significant error due to the distortion when the incident angle becomes small.
The circle fitting on the hand motion (green) still has the error in position.
After the refinement, the articulation parameters (red) have been clearly improved.

\subsection{Other results}
Figure~\ref{fig:others} shows the results of several scenes, including both prismatic or revolute objects.
Only for the prismatic object in the top-right in Fig.~\ref{fig:others}, we applied the hand region detector \cite{Dibia2017} since there was not enough body area in the picture for detecting hand positions by OpenPose.
One of the advantages of the proposed method is robustness.
In the initial estimation by the hand motion, the proposed method correctly identified the articulation type of the objects.
In all scenes of Fig.~\ref{fig:others}, the final segmentation results demonstrate that our method correctly estimated the manipulated object's region without ambiguous regions.

\section{Conclusion}
In this paper, we presented a hand-motion-guided articulation model estimation and segmentation method.
The proposed method uses the hand position information for the initial articulation estimation, the model refinement by a constrained ICP alignment, and selecting the region of the manipulated object from ambiguous regions.
The experimental results demonstrate that our method correctly estimates the articulation model and segmentation of manipulated objects.
For future work, we will extend the method and apply it for articulated objects with more degrees of freedom.

{\small
	\bibliographystyle{IEEEtran}
	\bibliography{myrefs}
}

\end{document}